Hyperlocal disaster damage assessment using bi-temporal street-view imagery and pre-trained vision models

Yifan Yang[a], Lei Zou[a]*, Bing Zhou[b,c], Daoyang Li[d], Binbin Lin[a], Joynal Abedin[a], Mingzheng Yang[a]

[a]Department of Geography, Texas A&M University, College Station, USA
[b]Department of Geography, Pennsylvania State University, University Park, USA
[c]Department of Geography and Sustainability, University of Tennessee, Knoxville
[d]Spatial Sciences Institute, University of Southern California, Los Angeles, USA
*Corresponding author: lzou@tamu.edu

**Abstract**

Street-view images offer unique advantages for disaster damage estimation as they capture impacts from a visual perspective and provide detailed, on-the-ground insights. Despite several investigations attempting to analyze street-view images for damage estimation, they mainly focus on post-disaster images. The potential of time-series street-view images remains underexplored. Pre-disaster images provide valuable benchmarks for accurate damage estimations at building and street levels. These images could aid annotators in objectively labeling post-disaster impacts, improving the reliability of labeled data sets for model training, and potentially enhancing the model performance in damage evaluation. The goal of this study is to estimate hyperlocal, on-the-ground disaster damages using bi-temporal street-view images and advanced pre-trained vision models. Street-view images before and after 2024 Hurricane Milton in Horseshoe Beach, Florida, were collected for experiments. The objectives are: (1) to assess the performance gains of incorporating pre-disaster street-view images as a no-damage category in fine-tuning pre-trained models, including Swin Transformer and ConvNeXt, for damage level classification; (2) to design and evaluate a dual-channel algorithm that reads pair-wise pre- and post-disaster street-view images for hyperlocal damage assessment. The results indicate that incorporating pre-disaster street-view images and employing a dual-channel processing framework can significantly enhance damage assessment accuracy. The accuracy improves from 66.14% with the Swin Transformer baseline to 77.11% with the dual-channel Feature-Fusion ConvNeXt model. Gradient-weighted Class Activation Mapping (Grad-CAM) shows that incorporating pre-disaster images improves the pre-trained vision model's capacity to focus on major changes between pre- and post-disaster images, thus enhancing model performance. This research enables rapid, operational damage assessments at hyperlocal spatial resolutions, providing valuable insights to support effective decision-making in disaster management and resilience planning.

# 1 Introduction

Timely and accurate disaster damage assessment, including the evaluation of impact types and intensities, is crucial for effective disaster management. It plays a pivotal role in identifying victims in need, locating damaged infrastructures, prioritizing emergency resource allocation, and enabling efficient recovery efforts. Traditional assessment methods, such as field surveys or satellite-based observations, often struggle to provide both extensive spatial coverage and hyperlocal details simultaneously. The growing availability of street-view imagery has unlocked a range of applications, including building-level disaster damage assessment. Street-view imagery captures disaster impacts from a human visual perspective and can deliver detailed, on-the-ground information, such as damages to windows, doors, and facades (Zhai & Peng, 2020). Such information is imperative in understanding disaster impacts, but its collection is difficult through aerial images and time-consuming through fieldwork. Contrarily, systematically obtaining street-view imagery throughout the post-disaster response and recovery phases is operationally feasible and scalable, enabling large-scale, hyperlocal, and efficient damage assessments and restoration monitoring across affected areas.

Assessing disaster losses using street-view imagery typically involves two key steps: image geolocation and damage evaluation. Recent Artificial Intelligence (AI) advances have introduced innovative approaches to enhance both processes. For example, a cross-view geolocation and disaster perception framework was developed to process street-view imagery, leading to significant improvements in the accuracy of disaster perception and geolocation for affected areas (Li et al., 2024). Furthermore, researchers have proposed a multimodal Swin Transformer model that integrates street-view imagery with structured supplementary data, such as building age and wind speed, to classify building damage following hurricanes (Xue et al., 2024). These previous efforts highlight the growing potential of combining cutting-edge AI techniques with street-view imagery to advance disaster damage assessment and improve resilience planning.

However, existing research has not yet fully explored the potential of time-series street-view imagery, i.e., combining pre- and post-disaster images, in hyperlocal damage assessment. Pre-disaster images could serve as a benchmark, capturing building and street conditions unaffected by disasters. Figure 1 illustrates a side-by-side comparison of street-view images obtained before and after a hurricane. The red rectangles in post-disaster images on the left show intact open spaces. However, by comparing them with the pre-disaster images on the right, it becomes evident that the hurricane has caused catastrophic impacts by removing buildings in these areas.

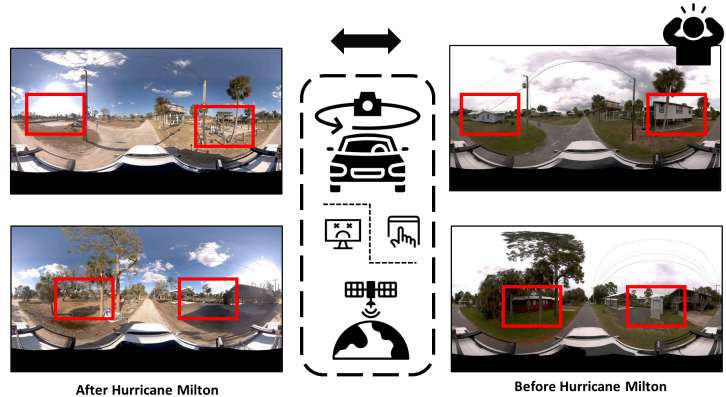

Figure 1. Pre- and Post-Hurricane Street View Comparison for Disaster Impact Assessment.

The above examples demonstrate the limitations of relying solely on post-disaster street-view imagery for damage assessment. It also reveals that designing and training AI models to analyze differences between pre- and post-disaster street-view images has the potential to create a more accurate and dynamic approach to assessing post-disaster losses. Additionally, annotators can objectively and consistently label disaster impacts by comparing these images, enhancing the reliability of labeled datasets. This, in turn, strengthens the training and fine-tuning of AI-based models, ultimately improving the accuracy of disaster damage assessments.

This investigation aims to fill the gap by building and evaluating a dual-channel neural network model to capture the information of bi-temporal street-view images for hyperlocal disaster estimation. Street-view images taken before and after the 2024 Hurricane Milton in Horseshoe Beach, Florida, U.S., were collected for experiments. The objectives are: (1) to evaluate the performance improvement of incorporating pre-disaster street-view imagery into fine-tuning pre-trained vision models (Swin Transformer and ConvNeXt) for damage estimation; (2) to design and evaluate a dual-channel neural network approach that reads pair-wise pre- and post-disaster street-view images for hyperlocal damage assessment. We hypothesize that leveraging bi-temporal street-view imagery and a dual-channel framework coupled with pre-trained vision models improves damage estimation accuracy. This research can support rapid street- and building-level damage estimations, providing information for effective decision-making in disaster response, recovery, and resilience planning. The produced open-source dataset can serve as a benchmark for computer vision and disaster research.

The article proceeds as follows. Section 2 provides a comprehensive review of work applying street-view imagery and vision models in disaster management. Section 3 begins by introducing Hurricane Milton, followed by the dataset collection and preparation. It also outlines the methodologies, including disaster estimation using pre-trained vision models, the integration of pre-disaster images, and the development of a dual-channel approach for bi-temporal analysis. Section 4 presents the experiment results, emphasizing the performance of pretrained models, the profit of bi-temporal data, and the effectiveness of the dual-channel model. Section 5 elaborates on the implications, critically examines the limitations, and proposes directions for future research. Finally, Section 6 concludes the paper, highlighting its scientific contributions and the broader applicability of pretrained vision models in advancing disaster resilience research and practice.

## 2 Related Work

### 2.1 Street-View Imagery for Disaster Impact Assessment

Street-view imagery offers unique advantages for studying disaster impacts by providing rich and detailed ground-level perspectives, complementing the limitations of traditional aerial and satellite imagery, thereby enhancing disaster perception and damage assessment (Zhai & Peng, 2020). The growing availability of street-view images in open platforms such as Mapillary, Google Street View, and OpenStreetMap (OSM) further enhance their applicability in research and practice (Mahabir et al., 2020; Neuhold et al., 2017), e.g., studying urban visual perception (Gao et al., 2025; Ito et al., 2024; Li et al., 2024; Yang et al., 2024).

Given the complex, unstructured, and multi-dimensional nature of information in street-view images, advanced deep learning frameworks are essential for accurate information extraction. One study assessed urban environmental visual quality with street-view imagery, employing the Pyramid Scene Parsing Network (PSPNet) for semantic segmentation to capture fine-grained urban scene elements (Sun et al., 2023). This research demonstrated the critical role of time-series imagery in evaluating spatial equity, referring to the fairness in the spatial distribution of perceived urban street qualities such as safety, wealth, and beauty. Another investigation introduced an innovative deep learning approach, "SegFormer-B5 + ConvNeXt-Base + RF," to clarify the relationship between objective visual elements—such as buildings, greenery, and sky—extracted from street-view imagery, and subjective human perceptions like safety, beauty, and liveliness (Zhao, Lu, et al., 2024). Meanwhile, a literature review analyzed 393 studies using street-view imagery with deep learning for urban visual perception analysis (Ito et al., 2024), highlighting major challenges in training data preparation, such as subjectivity in human labeling, annotation inconsistency, and low-quality image datasets. These issues underscore the need for improved data curation and more robust labeling protocols.

The integration of street-view imagery with deep learning frameworks has been effectively applied in disaster damage assessment. A recent study evaluated the performance of various pre-trained image classification models in assessing post-hurricane building damage (Xue et al., 2024). They proposed the Multi-Modal Swin Transformer (MMST), a novel framework that combines imagery with structured data such as building value, building age, and wind speed. Trained on data from Hurricane Ian in Florida (2022), MMST achieved an accuracy of 92.67%, outperforming other state-of-the-art models by 7.71%. Disaster street-view imagery can also be effectively integrated with remote sensing data. Li et al. (2024) developed a cross-view disaster mapping framework called CVDisaster, which fuses street-view imagery and very high-resolution satellite imagery to enable simultaneous geolocation and damage perception. Using the 2022 Hurricane Ian as a case study, this investigation introduced a novel cross-view dataset (CVIAN) and demonstrated that CVDisaster achieved over 80% accuracy in geolocation and 75% in damage perception with limited data for fine-tuning. This highlights the potential of cross-view learning to enhance situational awareness and perception-based assessments in post-disaster scenarios. Moreover, C. Wang et al. (2024) investigated the relationship between household vulnerability and post-disaster street-view imagery of buildings. Their pilot study demonstrates a strong correlation between image-based predictions and vulnerability indicators, suggesting that street-view imagery could serve as a cost-effective supplement for large-scale vulnerability assessments.

While research on post-disaster street-view imagery has grown extensively, many studies have underscored a notable gap in examining the temporal aspects of street-view data during disaster events. Specifically, current literature highlights a lack of research exploring the combined use of pre- and post-disaster imagery, partially due to data unavailability. There remains significant room for investigation into how temporal street view imagery, spanning different timeframes, impacts disaster perception and damage assessment models. This represents a critical area for further exploration to enhance our understanding of disaster dynamics (de Alwis Pitts & So, 2017; Li et al., 2024; Mabon, 2016).

## 2.2 Pre-Trained Vision Models for Disaster Perception

With the continuous advancement of deep learning technology, fine-tuning pre-trained models has become a common approach to achieve satisfactory results (Han et al., 2021; Zong et al., 2024). Early studies primarily utilized pre-trained language models for tasks involving disaster-related textual data (Koshy & Elango, 2023; Z. Liu et al., 2021; Maswadi et al., 2024; Pavani & Malla, 2024). With the increasing use of street view imagery datasets in disaster perception tasks, researchers have shifted towards employing vision-based pre-trained models, e.g., Residual Network (ResNet), Vision Transformer (ViT), Visual Geometry Group (VGG), for disaster perception and damage classification tasks (Mahabir et al., 2020; C. Wang et al., 2024; Xue et al., 2024; Zhai & Peng, 2020; Zhao, Lu, et al., 2024). Compared to text-based methods, this approach avoids potential information loss during textual descriptions by directly extracting features from images. Concurrently, challenges exist in image-based research, such as the inherent subjectivity in annotated image datasets and the less intuitive visualization of classification results compared to textual data.

Among vision-based pre-trained models, the Swin Transformer and ConvNeXt have emerged as two of the most promising architectures for disaster perception tasks using street-view images. The Swin Transformer is a hierarchical vision Transformer that introduces a shifted window-based self-attention mechanism, allowing for efficient computation and effective modeling of both local and global dependencies (Z. Liu et al., 2021; Zhang et al., 2022). In contrast, ConvNeXt is a modernized convolutional neural network (CNN) architecture that revisits and refines standard ResNet-like designs with state-of-the-art techniques (Woo et al., 2023; Zhao, Ma, et al., 2024). ConvNeXt balances simplicity and performance, making it highly adaptable to disaster-related classification tasks, especially in scenarios with limited computational resources or training data. Drawing from the strengths of these architectures, this study focuses on Swin Transformer and ConvNeXt as representative models for evaluating the value of incorporating pre-disaster images in disaster perception. Their distinct architectural principles—Transformer-based attention versus convolutional design—provide a robust and comparative foundation to illustrate how pre-disaster imagery influences perception accuracy and interpretability.

Standing on top of the success of singular models, the introduction of dual-channel architecture, which usually incorporates multiple models to dissect time series data, creates a novel avenue to process multi-temporal data, such as pre-disaster and post-disaster remote sensing images. Each channel within the architecture extracts key features at different timestamps (Asokan et al., 2020), and the feature fusion elements in the architecture combine information from these channels to create a comprehensive representation, enhancing the model's ability to detect temporal changes (Asokan et al., 2020; Luo et al., 2023; Quan et al., 2022; Tao et al., 2024). The cross-attention mechanism further refines this process by allowing the model to focus on critical regions of change between the two sets of images (F. Liu et al., 2024; L. Wang et al., 2024). This study adopted the dual-channel architectures, feature fusion techniques, and cross-attention mechanisms and applied them to process pre- and post-disaster street-view images.

## 3 Data and Methodology

### 3.1 Data Collection and Preparation

Hurricane Milton, a highly destructive tropical cyclone, was formed in October 2024 and made landfall along Florida's western coastline on October 9. It caused significant damage throughout the state of Florida. According to the National Centers for Environmental Information (NCEI) under the National Oceanic and Atmospheric Administration (NOAA), Milton resulted in an estimated economic loss of $34.3 billion and claimed the lives of at least 35 people (Smith, 2020). Horseshoe Beach is a town in southern Dixie County, Florida, United States, which suffered extensive damage from Hurricane Milton. We selected Horseshoe Beach as the study area because comprehensive, high-quality street-view data captured both before and after the disaster are available in this region. The track of Hurricane Milton, along with the location of Horseshoe beach, are displayed in Figure 2.

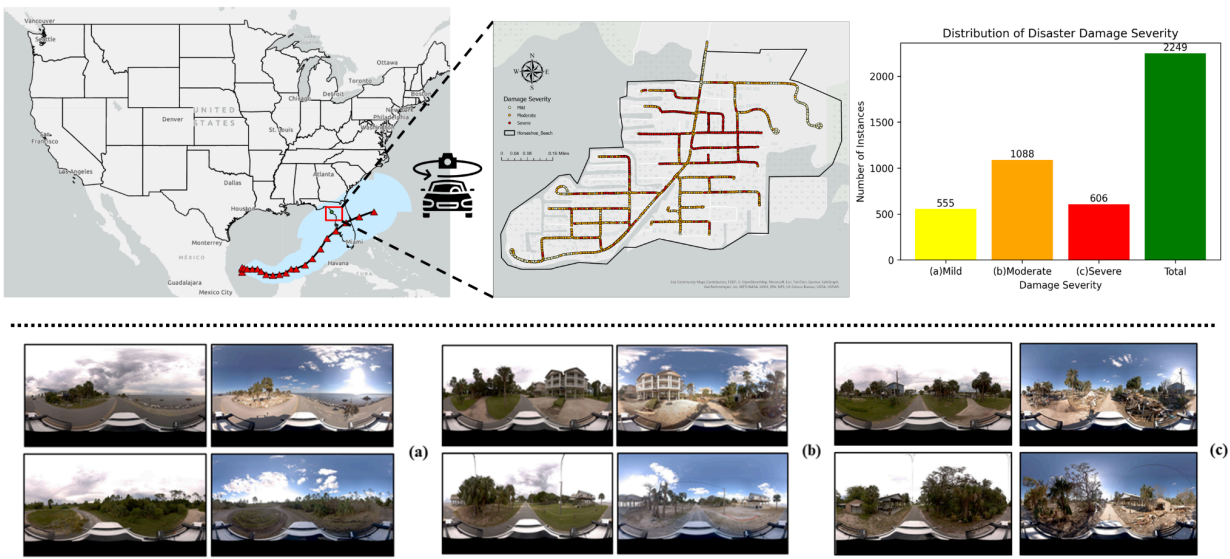

Figure 2. Study area and samples of street-view image pairs in Horseshoe Beach area of Florida.

The pre- and post-Milton street-view data were collected from Mapillary (https://www.mapillary.com/), an open-source platform for street-level image and map data collection, management, visualization, and sharing. A total of 2,610 street-view images on August 29, 2023 (pre-Milton phase) and 3,246 images on October 17, 2024 (post-Milton phase) were obtained from Mapillary. These images were captured by Site Tour 360. Latitude and longitude metadata were used to match post-disaster images with the nearest pre-disaster images. After filtering out unmatched instances, 2249 pairs of street-view images were obtained.

The next step is manual annotation. We manually labeled post-disaster images into three categories, mild, moderate, and severe, based on damage conditions. To establish clear classification criteria for disaster impacts, we focused on four primary features: fallen trees, building debris, destroyed infrastructures, and inundated roads. Images labeled as "Mild" are characterized by clean scenes with no major property damage or only minor disruptions, such as small areas of fallen trees, and are considered to have caused negligible economic loss. "Moderate" images exhibit more impacts compared to "Mild" ones, typically featuring more pronounced fallen trees, some building debris, and visible pools of water around the base of trees. These images indicate noticeable economic damage, though not to the extent of severe destruction. "Severe" images represent the most chaotic cases, distinguished by extensive or widespread tree falls, flooded roads, and significant building debris. These scenarios are identified as causing substantial property loss. Figure 2 (a) – (c) illustrates examples of pair-wise street-view images falling into the three damage severity levels. The dataset annotations were finalized through a cross-validation process involving four annotators, ensuring consistency and accuracy. After manual annotation, 555 images were labeled as "Mild", 1,088 as "Moderate", and 606 as "Severe".

## 3.2 Experimental Design

Three distinct experiments were designed, as illustrated in Figure 3. This study employed four pre-trained vision models: ConvNeXt-Tiny, ConvNeXt-Base, Swin Small Patch Window7 224, and Swin Base Patch Window7 224. The first experiment used pre-trained models with post-disaster images as input, classifying them into three damage categories. The second experiment introduced pre-disaster images as a no-damage baseline, using pre-trained models to classify images into four categories: no damage, mild, moderate, and severe damage. Both experiments consist of three steps: (1) data preprocessing (e.g., standardization and normalization), (2) feature extraction using fine-tuned pre-trained models, and (3) damage classification.

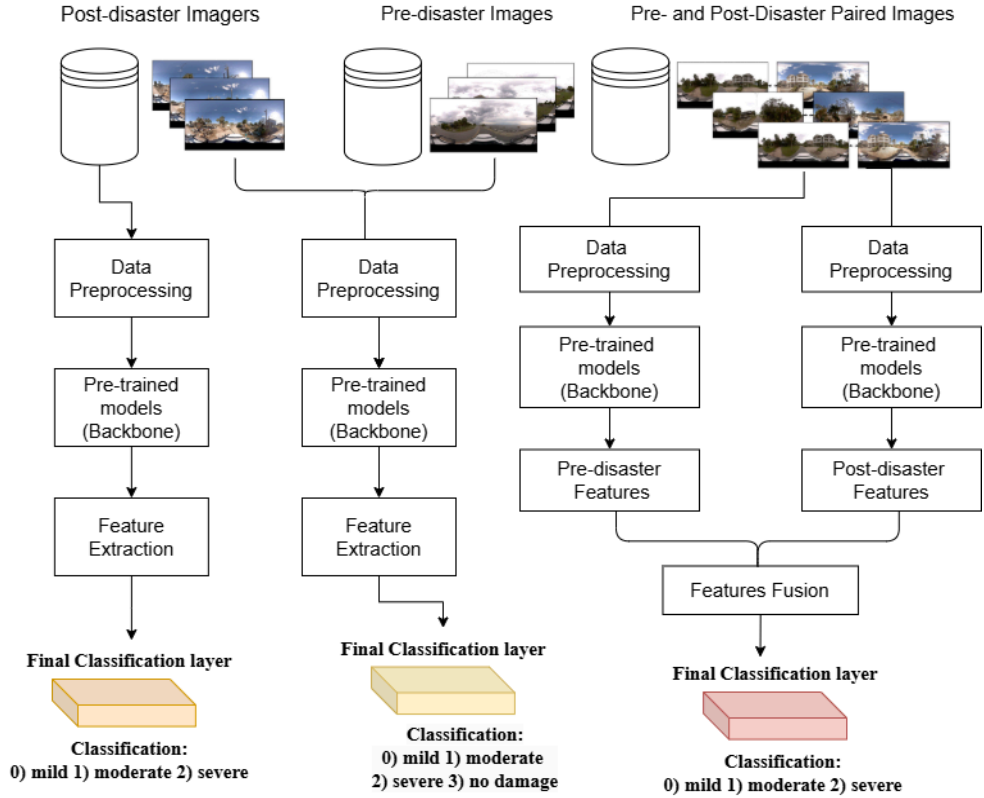

Figure 3. Comparison of Single-Phase and Bi-Temporal Disaster Classification

Simply treating pre-disaster images as a fourth class (no damage) to fine-tune pre-trained vision models has certain limitations. This approach overlooks the temporal continuity of disaster progression, treating pre-disaster images as static categories rather than dynamic reference points. As a result, the model may struggle to capture the underlying relationships between pre-disaster and post-disaster images, limiting its ability to effectively assess damage. Therefore, in the third experiment, we designed a dual-channel model with pairwise pre- and post-disaster street-view images as input, categorizing damage into three levels. Initially, features were independently extracted from pre- and post-disaster images following the same three steps in the first two experiments. These extracted features were fused using multiple strategies, i.e., feature fusion, cross-attention mechanisms, and Siamese network-based approaches, to quantify disaster-induced changes from bi-temporal images.

All three experiments were conducted in a GPU-accelerated environment using the PyTorch deep learning framework for model training and inference. To ensure robust model performance, the dataset was split into training and validation sets with varying ratios: 9:1, 8:2, 7:3, 6:4, 5:5, and 4:6, to evaluate generalizability across data distributions. Data preprocessing steps include image resizing to 224×224 pixels, normalization to match the input distribution of pre-trained models, and several data augmentation techniques—such as random cropping, horizontal flipping, and color jittering—to improve generalization. For optimization, the AdamW optimizer was used in conjunction with a cosine annealing learning rate scheduler. The models were trained with the cross-entropy loss function for this multi-category classification task. Once the images are fed into the model, hierarchical features are extracted through either the ConvNeXt or Swin Transformer backbone, followed by a fully connected layer to produce the final classification outputs corresponding to the three or four damage levels. The technical details of the pre-trained models for feature extraction and fusion in the three experiments are elaborated in the subsequent sections 3.3-3.5.

We utilized four evaluation metrics, precision (P), recall (R), F1 score, and category-specific recall, to comprehensively assess model performance. A True Positive (TP) refers to a street-view image that was correctly classified into its respective damage level. A False Positive (FP) for a class occurs when the model incorrectly assigns an image to this class, while a False Negative (FN) for a class represents an image belonging to this class is overlooked. Precision measures the proportion of correctly classified samples (TP) among all predictions for a given class (TP + FP) (Equation 1), and recall quantifies the proportion of correctly identified positive samples (TP) relative to the total actual positives (TP + FN) (Equation 2). The F1-score, calculated as the harmonic mean of precision and recall, provides a balanced metric to evaluate the model's accuracy and robustness (Equation 3). In this study, we report weighted-average precision, recall, and F1 scores across all classes to account for class imbalance and provide an overall performance summary. Rather than computing per-class metrics, this averaging approach reflects the contribution of each class based on its support (number of true instances). To standardize disaster classification, all post-disaster images were grouped into three predefined damage severity levels: Class 0 (mild), Class 1 (moderate), and Class 2 (severe). In addition, we report category-specific recall, which calculates the proportion of correctly classified images within each class relative to the total number of samples in that class. This metric offers a more granular view of model performance across damage levels.

$$Precision = \frac{TP}{TP+FP} \tag{1}$$

$$Recall = \frac{TP}{TP+FN} \tag{2}$$

$$F1 - score = 2 * \frac{Precision*Recall}{Precision+Recall} \tag{3}$$

To enhance the interpretability of tested models and gain insight into how they identify disaster-related damage from post-disaster street-view imagery, we employed Gradient Weighted Class Activation Mapping (Grad-CAM). Grad-CAM generates heatmaps to highlight the most influential regions in the image that contribute to the model's classification decisions. It uses the gradients of the target class flowing into the final convolutional layer to produce a coarse localization map, indicating the important regions in the input street-view imagery. By applying Grad-CAM, we can visually identify parts of post-disaster images critical for assessing damage severity. This technique allows us to compare attention patterns across the three approaches.

### 3.3 Pre-Trained Vision Models with Post-Disaster Images

The first experiment explores the applicability and performance of different pre-trained vision models for disaster damage classification solely based on post-disaster street-view imagery. Figure 4 illustrates the framework of the first experiment with ConvNeXt and Swin Transformer models.

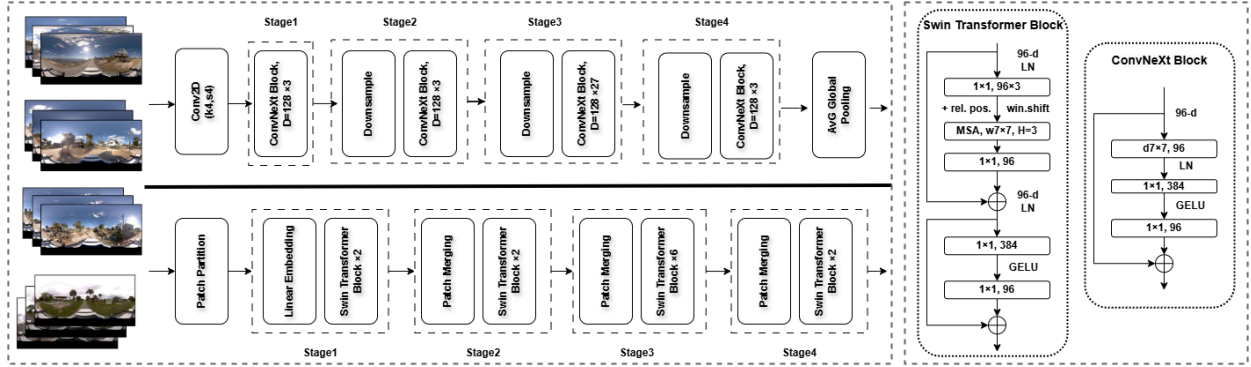

Figure 4. ConvNeXt and Swin Transformer for multi-stage feature extraction architecture of disaster perception.

The ConvNeXt series is a modernized architecture based on Convolutional Neural Networks (CNNs), designed with optimized components to enhance image feature extraction (Woo et al., 2023; Yu et al., 2024; Zheng, 2024). ConvNeXt models consist of four stages, where each stage employs convolutional layers (ConvNeXt Blocks) and downsampling operations to progressively reduce image resolution while improving the semantic richness of the extracted features. Each ConvNeXt Block utilizes large 7*7 convolutional kernels, linear normalization, and a Gaussian Error Linear Unit (GELU) activation function. The convolutional kernels help effectively capture spatial features over broader regions. The GELU activation function further enhances the non-linear representation of features while mitigating issues such as gradient explosion or vanishing gradients. ConvNeXt-Tiny is tailored for scenarios with limited computational resources, while ConvNeXt-Base provides more robust feature representation capabilities. ConvNeXt models take advantage of the simplicity and efficiency of CNN and focuses on spatial feature extraction.

The Swin Transformer series is a multi-head self-attention (MSA)-based model designed for processing high-resolution images. This model first leverages Patch Partition to divide input images into fixed-size patches (7x7) to reduce computational complexity while preserving local spatial information. Then the model employs a four-stage workflow with each stage containing a Swin Transformer block. In each block, the multi-head self-attention mechanism (MSA) captures long-range dependencies in the data, making it highly effective in modeling the widespread damage patterns observed in disaster scenarios. The Shifted Window Attention mechanism builds upon standard window-based self-attention by introducing a window-shifting strategy between successive transformer layers, which enables cross-window connections and enhances global context modeling (Z. Liu et al., 2021, 2022; Xie et al., 2021). While traditional window-based self-attention restricts information flow within local non-overlapping windows, the shifted window approach allows information from adjacent windows to interact by shifting the window partitioning pattern. This reduces the model's reliance on specific spatial regions and improves its ability to capture long-range dependencies and holistic structural patterns across the entire image. Swin Small is a lightweight version optimized for efficient computation, while Swin Base offers enhanced feature representation capabilities. Swin Transformers excels at capturing both global and local relationships, making it an ideal choice for complex disaster images where dependencies span large areas.

## 3.4 Pre-Trained Vision Models with Bi-Temporal Images

The second experiment incorporates pre-disaster street-view imagery as supplementary data along with post-disaster images to fine-tune pre-trained vision models. To accommodate the introduction of pre-disaster data, the models were fine-tuned to classify each image into one of four categories: no damage (pre-disaster imagery), mild, moderate, and severe damage. By integrating pre-disaster imagery as an additional class, the model benefits from the added context, particularly in comprehensively detecting disaster impacts across the three damage categories. To evaluate the performance of models incorporating pre-disaster imagery, we focused on calculating metrics for the three damage categories: mild, moderate, and severe, as the inclusion of pre-disaster imagery as a fourth class ("no damage") serves only as a reference point. This evaluation strategy ensures that the addition of pre-disaster imagery enhances disaster classification performance without artificially inflating metrics by incorporating "no damage" as an auxiliary class.

## 3.5 Dual-Channel Models with Pairwise Bi-Temporal Images

To enable the model to compare pre- and post-disaster information at the same location, the third experiment proposes a dual-channel model framework. As illustrated in Figure 5, the dual-channel model processes pairs of pre-disaster and post-disaster images as independent inputs, utilizing pre-trained models for feature extraction. We utilized pre-trained ConvNeXt and Swin Transformer backbones as fixed feature extractors. By freezing their parameters during training, we preserved their learned visual representations and focused on training fusion layers and the classifier. This approach reduces training complexity and prevents overfitting, especially when working with limited training datasets. To effectively capture disaster-related changes, we explored five dual-channel processing strategies (S1-S5 in Figure 5), totaling seven models.

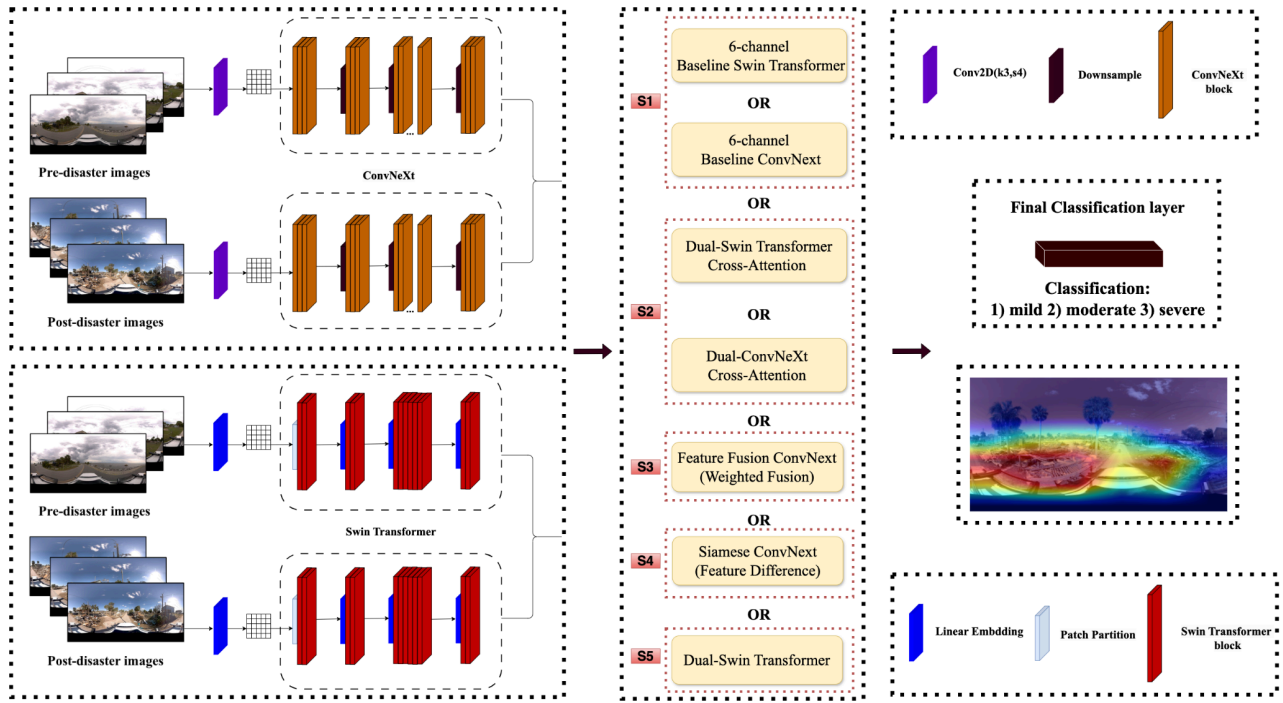

Figure 5. Dual-Channel training process for Bi-Temporal SVI Disaster Perception.

The first strategy concatenates the raw pre-disaster and post-disaster street-view images along the channel dimension, resulting in a single 6-channel input (i.e., stacking two 3-channel RGB images). To accommodate this format, the input layer of the pre-trained Swin Transformer and ConvNeXt models was modified to accept 6-channel inputs. The concatenated input is then processed through the selected backbone (either Swin Transformer or ConvNeXt) to extract joint features, which are subsequently fed into a classifier for damage level prediction. This approach serves as the baseline model, leveraging the prominent feature extraction capabilities of the backbones while maintaining a relatively simple architecture. However, it does not model the temporal or semantic interactions between the pre- and post-disaster images. This baseline model can assess the discriminative power of dual-channel pre-trained vision models without explicit change detections.

The second strategy leverages the Dual-Swin Transformer (ConvNeXt) Cross-Attention Channel. This method enhances the interactions between pre-disaster and post-disaster image features using a cross-attention mechanism. Instead of simply concatenating features, it models the relationship between them by computing the attention between encoded images in the two timestamps. In this approach, pre-disaster features $F_{pre}$ act as queries (Q), while post-disaster features $F_{post}$ act as both keys (K) and values (V) in the attention mechanism (Equations 4-6). $W_q$, $W_k$, $W_v$ are learnable parameter matrices that convert features into inputs to attention mechanisms. This enables the model to dynamically focus on relevant regions in post-disaster images based on the pre-disaster context.

$$Q = W_q F_{pre}, \; K = W_k F_{post}, \; V = W_v F_{post} \qquad (4)$$

$$Attention(Q, K, V) = softmax(\frac{QK^T}{\sqrt{d_x}})V \qquad (5)$$

$$F_{cross} = Attention(F_{pre}, F_{post}, F_{post}) \qquad (6)$$

The third strategy is based on feature fusion ConvNeXt. ConvNeXt is suitable for weighted fusion methods due to its convolution-based local feature modeling, linear structure adaptability, and computational stability. In contrast, the Swin Transformer is less suitable for feature fusion methods due to the non-uniform nature of window-based attention, its emphasis on global dependency modeling, and the impact of hierarchical feature representations. Hence, we chose feature fusion combined with ConvNeXt. This method fuses features extracted from the pre-disaster and post-disaster stages through an adaptive weighting mechanism. It calculates the linear weighted sum of the pre-disaster and post-disaster features (Equation 7):

$$F_{fused} = \alpha F_{pre} + (1-\alpha) F_{post} \qquad (7)$$

where $F_{pre}$ and $F_{post}$ are feature vectors extracted from the pre- and post-disaster images, respectively, $F_{fused}$ is the final fused representation passed to the classifier, and α is a learnable parameter that represents the importance of the pre-disaster information. By learning the optimal weighting method, the model can automatically adjust the contribution of pre-disaster and post-disaster information according to different damage degrees. Finally, the fused features are passed to the classifier for final prediction.

The fourth strategy is Siamese ConvNeXt, which performs feature difference calculation. Unlike the feature fusion approach, the Siamese strategy focuses on capturing disaster-induced changes by computing the absolute difference between the two feature sets. Specifically, ConvNeXt is used to extract features from the pre-disaster and post-disaster images independently. The model then computes their absolute differences to emphasize regions affected by the disaster. These differences in features are subsequently fed into a classifier for damage prediction. We chose ConvNeXt because its convolutional architecture makes it well-suited for this difference-based approach. In contrast, the non-local and window-based attention mechanism in Swin Transformer may introduce challenges in aligning fine-grained features across bi-temporal inputs, which can affect the precision of difference calculations. This strategy is particularly effective for datasets exhibiting significant structural damage, where the contrast between pre- and post-disaster imagery is more pronounced. However, it may perform less effectively when changes are subtle or visually ambiguous.

The last strategy is Dual-Swin Transformer, which uses two independent Swin Transformers to process pre- and post-disaster images respectively to extract spatiotemporal features and fuse them. Specifically, each image is first processed through a separate Swin Transformer and then fused in the high-level feature space through an attention mechanism (Equations 8-9):

$$F_{pre} = SwinTransformer(I_{pre}) , F_{post} = SwinTransformer(I_{post}) \qquad (8)$$

$$F_{concat} = Attention(F_{pre}, F_{post}) \qquad (9)$$

Finally, the fused features are input into the classifier for damage prediction. ConvNeXt is effective in local feature extraction but may have limitations in modeling large-scale changes and complex interactions. Therefore, we used the Swin Transformer for dual-channel experiments in this strategy. This method effectively leverages bi-temporal imagery to provide a robust and scalable solution for disaster perception and classification. By integrating temporal context and utilizing advanced feature fusion techniques, the dual-channel model framework has the potential to significantly improve the accuracy and interpretability of disaster analysis.

## 4 Results

### 4.1 Pre-Trained Models with Post-Disaster Images

In all experiments, we tested different splits of training and validation sets based on pre-trained vision models and post-disaster images. Most models achieve optimal performance when the training and validation split is 8:2. To ensure the consistency and comparability of results across different model settings, we fixed the training-validation split ratio at 8:2 for all subsequent experiments. The results of experiment one based on the 8:2 split are summarized in Table 1. ConvNeXt_tiny achieved the highest overall accuracy at 72.15%, which was 1.78% higher than Swin_base (70.37%) and 2.15% higher than Swin_small (70%). It also attained the best precision (0.7313), recall (0.7315), and F1 score (0.7179) among all models, surpassing the next-best F1 score (0.7022 for ConvNeXt_base) by 1.57%. The possible reason is that ConvNeXt_tiny, despite having fewer parameters, generalizes better in a data-limited setting due to its efficient feature extraction capabilities. Conversely, ConvNeXt_base, with a larger parameter set, might be overfit to specific patterns, leading to a slight drop in performance.

Table1 Performance of Pre-Trained Models with Post-Disaster Images

| Model name | Accuracy | P | R | F1 | Class-Specific Recall | | |
|---|---|---|---|---|---|---|---|
| | | | | | 0-mild | 1-moderate | 2-severe |
| **ConvNeXt_tiny** | **72.15%** | **0.7313** | **0.7315** | **0.7179** | **55.0%** | 82.37% | **68.82%** |
| ConvNeXt_base | 70.22% | 0.7256 | 0.7022 | 0.6961 | 50.93% | 85.19% | 61.05% |
| swin_small_patch_window7_224 | 70.00% | 0.7301 | 0.70 | 0.6879 | 44.04% | **88.84%** | 58.12% |
| swin_base_patch_window7_224 | 70.37% | 0.714 | 0.7037 | 0.6954 | 46.88% | 84.07% | 65.34% |

Per-class accuracies show that ConvNeXt_tiny outperformed other models in category 0 (mild) and category 2 (severe). In the mild category, ConvNeXt_tiny had a recall of 55.0%, which was 4.07% higher than ConvNeXt_base (50.93%), 10.96% higher than Swin_small (44.04%), and 8.12% higher than Swin_base (46.88%). In the severe category, ConvNeXt_tiny achieved the highest recall of 68.82%, exceeding Swin_base (65.34%) by 3.48% and Swin_small (58.12%) by 7.77%. It is worth noting that while the Swin models performed poorly in recognizing mild damage images, they accomplished satisfactory accuracies in recognizing moderate and severe damage images. Specifically, Swin_small achieved the highest recall of 88.84% in the moderate level, which was 6.47% higher than ConvNeXt_tiny's 82.37%. The results indicate that the Swin-based architecture may be more effective in capturing significant structural changes, particularly in cases of moderate and severe damage. To validate this observation, we presented Grad-CAM visualizations in the Section 4.4 to illustrate the model's attention to key structural features.

### 4.2 Pre-Trained Models with Bi-Temporal Images

Experiment two evaluates the performance of pre-trained vision models fine-tuned by both pre- and post-disaster imagery. The results are presented in Table 2. The models achieved near-perfect accuracy in classifying the fourth class (no damage). This section primarily elucidates accuracy improvements across the three damage classes. It should be noted that the reported accuracy accounts for all four categories, including the no damage class.

Table2 Performance of Pre-Trained Models with Bi-Temporal Data

| Model name | Accuracy | P | R | F1 | Class-Specific Recall |
|---|---|---|---|---|---|

|  |  |  |  |  | 0-mild | 1-moderate | 2-severe |
|---|---|---|---|---|---|---|---|
| **ConvNeXt_tiny** | 86.42% | 0.8698 | 0.8642 | 0.8617 | 46.41% | 81.71% | 71.50% |
| **ConvNeXt_base** | 87.32% | 0.8785 | 0.8732 | 0.87 | 47.62% | 85.27% | 71.90% |
| **swin_small_patch_window7_224** | 87.94% | 0.8924 | 0.8794 | 0.8732 | 40.68% | *91.82%* | *75.86%* |
| **swin_base_patch_window7_224** | *88.98%* | *0.8967* | *0.8898* | *0.8876* | *54.35%* | 87.62% | 67.24% |

Adding bi-temporal data significantly benefits Swin-based models, especially for moderate and severe damage classifications. Swin_small_patch_window7_224 achieved an outstanding recall of 91.82% in moderate damage identification, outperforming all other models. Likewise, Swin_base_patch_window7_224 achieved 87.62% in recall, marking an increase of 3.55% over the recall in experiment one (84.07%). The ConvNeXt models exhibited marginal improvements, with ConvNeXt_base improving from 85.19% to 85.27% and ConvNeXt_tiny decreasing from 82.37% to 81.71%. Concurrently, all models showed improvements in detecting images labeled as severe damage - Swin_small_patch_window7_224 had the highest recall of 75.86%, an improvement of 17.74% over the recall in experiment one. These results suggest that adding pre-disaster imagery enables the window-based self-attention mechanism in the Swin models to capture fine-grained local variations, making them particularly suitable at distinguishing moderate damages.

Despite the overall performance improvement, classifying mild damage images remains challenging for all models in Experiment two. The highest recall for mild damage was 54.35%, achieved by Swin_base_patch_window7_224, 7.47% higher than its performance in experiment one (46.88%). The recall of other models in this class was relatively low, with ConvNeXt_tiny at 46.41%, ConvNeXt_base at 47.62%, and Swin_small_patch_window7_224 at 40.68%.

### 4.3 Dual-Channel Models with Bi-Temporal Images

In the dual-channel models, pre-disaster and post-disaster images were input separately through two channels to enhance the models' ability to perceive and classify post-disaster imagery. Table 3 and Figure 6 present a comparative evaluation of different dual-channel models for disaster image classification. Table 3 and Figure 6(a) illustrate the overall accuracy of the seven tested models. Among them, the Feature-Fusion ConvNeXt achieved the highest validation accuracy at 77.11%, outperforming models relying solely on post-disaster images. The Dual-ConvNeXt + Cross-Attention and Dual-Swin + Cross-Attention models also exhibited prominence overall accuracies (76.91%), demonstrating the effectiveness of attention mechanisms in improving classification capabilities. Notably, these models performed better than baseline ConvNeXt (72.80%) and baseline Swin Transformer (66.14%), indicating that integrating cross-attention and feature fusion strategies with pre-trained vision models enhanced disaster perception.

Table 3. Performance of Dual-Channel Models with Bi-Temporal Images

| Model name | Accuracy | P | R | F1 | Class-Specific Recall | | |
| --- | --- | --- | --- | --- | --- | --- | --- |
| | | | | | 0-mild | 1-moderate | 2-severe |
| Baseline-ConvNext | 72.80% | 0.7315 | 0.7280 | 0.7256 | 56.91% | 79.12% | 75.54% |
| Baseline-Swin Transformer | 66.14% | 0.6889 | 0.6614 | 0.6481 | 39.02% | 84.34% | 57.55% |
| Dual-ConvNeXt+ Cross-Attention | 76.91% | 0.7704 | 0.7691 | 0.7672 | ***71.08%*** | 72.36% | ***91.37%*** |
| Dual-Swin Transformer + Cross-Attention | 76.91% | ***0.7970*** | 0.7691 | 0.7615 | 51.47% | ***91.97%*** | 74.60% |
| Dual-Swin Transformer | 74.17% | 0.7420 | 0.7417 | 0.7408 | 62.69% | 75.45% | 81.33% |
| Feature-fusion ConvNext | ***77.11%*** | 0.7751 | ***0.7710*** | ***0.7678*** | 59.39% | 82.72% | 84.44% |
| Siamese ConvNeXt | 75.54% | 0.7878 | 0.7554 | 0.7498 | 53.28% | 90.79% | 71.11% |

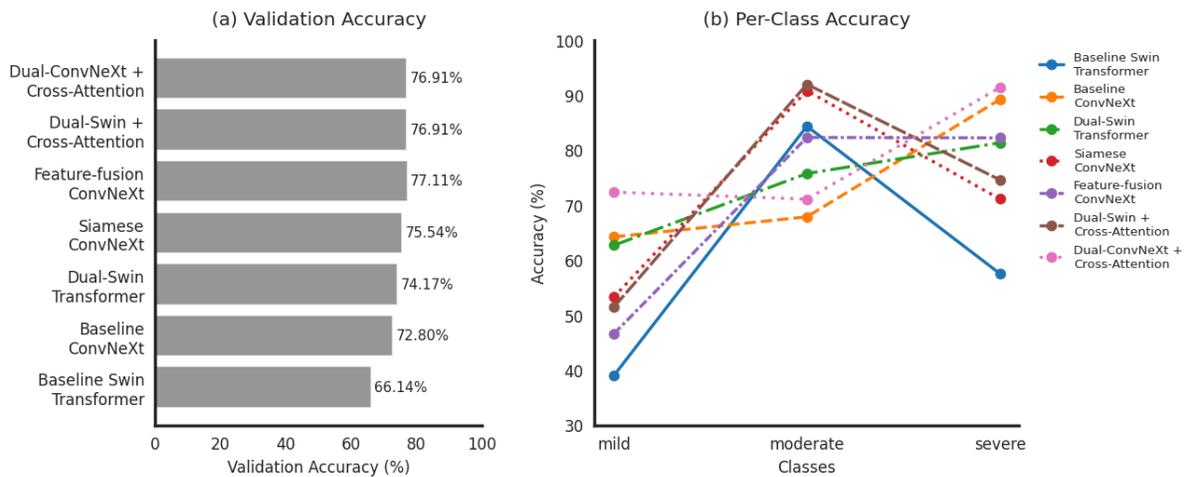

Figure 6. Comparison of dual-channel model performance on disaster damage classification tasks based on street-view images.

Table 3 and Figure 6(b) provide a detailed comparison of classification accuracy across three damage levels. The results indicate that models generally performed better on moderate (Class 1) and severe (Class 2) damage detections compared to mild damage identifications from street-view images. This suggests that larger-scale structural changes are easier to identify, whereas detecting subtle damages may be more challenging. The Dual-ConvNeXt + Cross-Attention and Dual-Swin + Cross-Attention models consistently show superior performance across all three categories, reinforcing the impact of the cross-attention strategy in refining damage classification. Additionally, feature fusion in ConvNeXt helps achieve robust predictions across different damage levels. These findings highlight the critical role of cross-attention and feature fusion mechanisms in enhancing disaster perception and classification accuracy using pre-trained vision models and bi-temporal street-view images.

### 4.4 Grad-CAM for Performance Evaluation

We conducted the Grad-CAM analysis and visualization based on untuned and fine-tuned pre-trained models for damage classification using post-disaster images to showcase Grad-CAM's capacity in interpreting model performance (Figure 7). The Grad-CAM visualizations correspond to the four fine-tuned pre-trained models evaluated in Experiment #1, which served as the initial baseline for post-disaster image classification. In each example, the first row Figure 7(1) displays the results from the untuned model, while the second row Figure 7(2) shows the corresponding results from the fine-tuned model. Heatmaps in each image highlight the model's focus areas, reflecting different models' perception of disaster-damaged areas. These regions help explain why certain models perform better in classifying disaster severity. We observe that fine-tuned models consistently highlight the most affected regions, such as collapsed structures and road obstructions, which aligns with human perception of disaster severity. This explains why fine-tuned models achieved higher classification accuracies and demonstrates the use of Gad-CAM in interpreting pre-trained models for decision-making. It should be noted that a quantitative evaluation of classification accuracy was not performed for the untuned models, as their performance was consistently suboptimal and not practically informative.

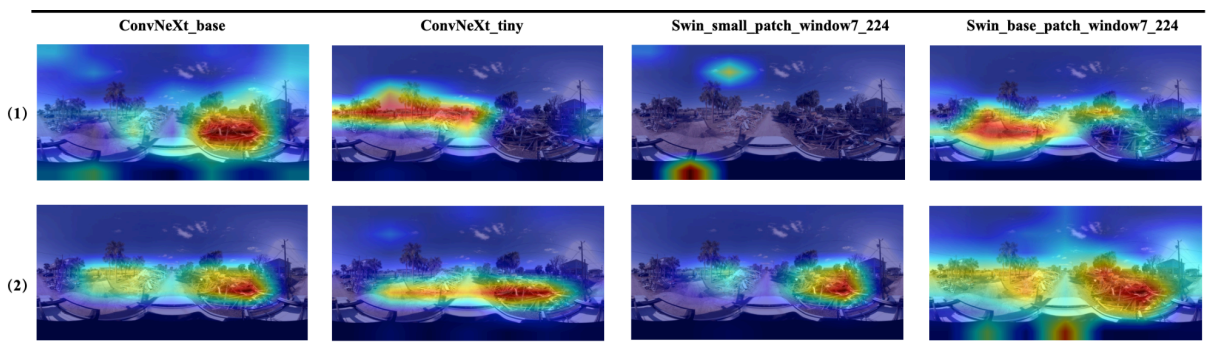

Figure 7. Grad-CAM Visualizations for Disaster Perception Using Untuned and Fine-Tuned Pre-Trained Models.

Figure 8 illustrates the Grad-CAM visualization results for three different post-disaster severity classes using the three experiments. We selected models that showed relatively stable performance in each experiment for visualization and interpretation. Panels (A) and (B) show three examples of the pre- and post-disaster street-view images with different damage levels. Panels (C) – (E) display the Grad-CAM visualizations for the model fine-tuned on post-disaster images, the model fine-tuned by a mixed dataset (both pre- and post-disaster images), and a dual-channel model fine-tuned by pairwise pre- and post-disaster images.

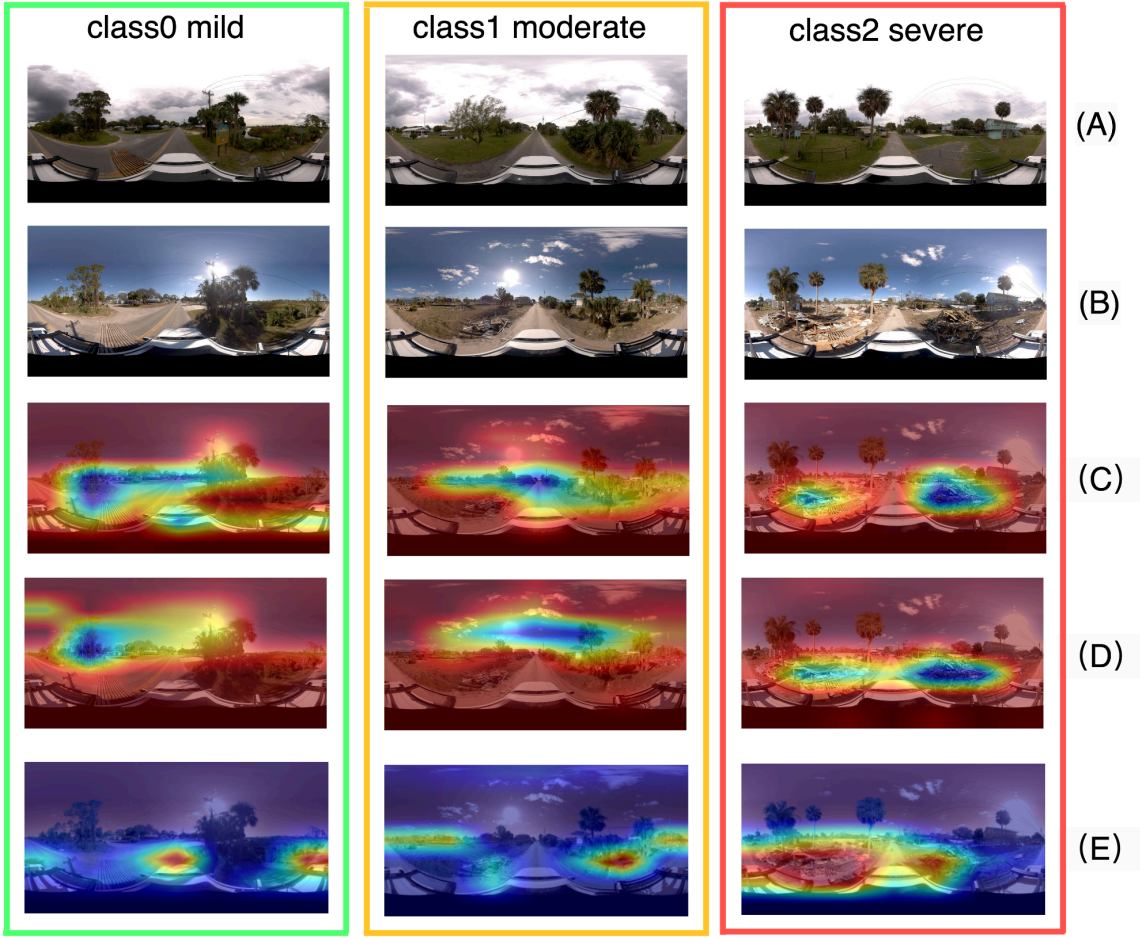

Figure 8. Results of Grad-CAM analysis of disaster perception using three different experiments.

Figure 8(C) reveals that since the optimal model in Experiment #1 (ConvNeXt_tiny) relies solely on post-disaster information, it tends to focus on damage features that are visually prominent in the aftermath (e.g., collapsed structures or obvious debris). However, it struggles to detect subtle or hidden damage, especially in the mild or moderate damage scenarios. As a result, the heatmap may activate irrelevant regions such as the sky or background. Furthermore, variations in lighting or camera angles in post-disaster images may cause the model to misplace its attention, leading to diffuse or overly broad activation zones.

Figure 8(D) shows that the Grad-CAM visualizations of the optimal model in Experiment #2 (swin_base_patch_window7_224). This model incorporates pre-disaster images during training, learns additional "normal" environmental features, and can therefore better highlight anomalies in the post-disaster scene. In mild cases, access to pre-disaster references allows the model to filter out unchanged areas and focus on actual damage, resulting in more precise and concentrated heatmap activation. However, in moderate scenarios, the model is influenced by external environmental factors such as lighting or weather, occasionally triggering excessive heatmap responses in irrelevant regions like the sky or distant backgrounds.

Figure 8(E) illustrates the performance of the optimal dual-channel model (Feature-Fusion ConvNeXt). This framework enables the model to directly compare images captured at the two time points, allowing it to focus on regions with the most significant changes. Through internal alignment and feature fusion, the model produces high-activation responses over actual damaged areas—such as collapsed buildings or fallen trees—while suppressing background noise. Compared to single-channel models or simple concatenation approaches, this dual-channel framework is more effective in capturing subtle but meaningful structural changes and more robust to variations in light conditions and camera angles.

## 5 Discussion

### 5.1 Significant Products, Findings, and Implications

This research has generated significant products, findings, and implications for designing, training, and applying pre-trained vision models for various applications, including disaster damage assessment. First, this study developed a unique dataset comprising bi-temporal (pre- and post-disaster) street-view images collected from Horseshoe Beach, Florida, which was significantly affected by Hurricane Milton in 2024. The dataset includes human-annotated perception labels to facilitate supervised learning. The availability of pre-disaster imagery is essential not only for model training but also for creating accurate manual annotations, as it provides a visual baseline for identifying and labeling changes in post-disaster images. This dataset addresses a critical gap in existing disaster perception research by enabling a structured evaluation of damage through bi-temporal visual comparisons.

Second, the experiments conducted using this dataset validated the importance of incorporating pre-disaster images in disaster perception classification tasks. Models that utilized both pre- and post-disaster street-view imagery consistently outperformed those relying solely on post-disaster inputs. This finding demonstrates that bi-temporal imagery enhances pre-trained vision models' ability to discern context-sensitive changes, particularly when damage is subtle or spatially complex. Notably, all models performed better when classifying moderate (Class 1) and severe (Class 2) damage, whereas identifying minor damage (Class 0) remained challenging. Despite this, the comparative nature of bi-temporal imagery has been proven valuable in improving classification performance across different severity levels.

To fully exploit the complementary nature of bi-temporal imagery, we evaluated a set of dual-channel deep learning architectures. Among these models, the Feature-Fusion ConvNeXt achieved the highest validation accuracy of 77.11%, significantly outperforming the Swin Transformer baseline, which achieved 66.14%. This reflects a notable accuracy improvement of 10.97% in damage assessment. Additionally, models utilizing cross-attention mechanisms, such as Dual-ConvNeXt + Cross-Attention and Dual-Swin + Cross-Attention, showed superior performance, further validating the effectiveness of modeling interactions between pre- and post-disaster features. In contrast, the Baseline Swin Transformer—which simply concatenated bi-temporal images—achieved the lowest accuracy, underscoring the limitations of naïve integration methods. These results highlight the critical role of architectural design in capturing meaningful spatiotemporal changes using multi-temporal information.

Beyond classification accuracy, this study also explored model interpretability through Grad-CAM visualizations. The visual outputs demonstrated that models like Feature-Fusion ConvNeXt could identify key damaged regions, such as collapsed buildings and blocked roads, while also detecting environmental changes like scattered debris and partially fallen trees. These interpretable visual explanations offer practical insights for real-world applications, especially in the context of rapid disaster response and recovery monitoring. The ability to accurately detect and explain damage patterns from bi-temporal street-view imagery positions this approach as a promising tool for near real-time disaster estimation and post-event assessment at the community level.

Beyond its scientific contributions, this study holds practical significance. The proposed approach is deployable and can be integrated into disaster response workflows used by government agencies, emergency response organizations, and humanitarian aid groups. The methodology's cost-effectiveness—where 360-degree cameras, street-view data, or drone imagery can be leveraged for rapid data collection—ensures that it can be widely implemented and scaled for real-world applications. Additionally, the labeled dataset generated in this study serves as a valuable resource for future research, supporting further advancements in disaster perception, multi-modal AI integration, and real-time disaster response optimization. By addressing existing challenges and exploring emerging technologies, future research can push the boundaries of disaster perception and resilience building through effective, data-driven strategies.

## 5.2 Limitations and Future Directions

Despite the significant findings, several limitations highlight opportunities for future improvements. One key consideration is the automation of the data annotation process using large language models (LLMs). A major challenge in disaster perception research using pre-trained models is the labor-intensive and time-consuming nature of manual data annotation. In this study, annotating 2249 pairs of pre- and post-disaster images required substantial human efforts, limiting the scalability of similar methodologies. To address this, automated annotation methods leveraging LLMs, such as GPT-4o-mini, could be explored. By integrating LLMs with semi-supervised learning techniques, annotation efficiency can be significantly improved while reducing human workload.

Figure 9 illustrates a preliminary exploration into the use of GPT-4o-mini for automatically annotating damage descriptions and assessments based on post-disaster street-view images. The results demonstrate that GPT-4o-mini can generate detailed textual descriptions of image content, alongside structured evaluations of disaster-related elements such as trees, buildings, debris, and water damage. These structured outputs can be visualized to enhance interpretability and facilitate large-scale dataset annotation. We also examined the drawing capabilities of GPT-4o to create post-disaster street-view images based on pre-disaster imagery and GPT-4o-mini-generated damage descriptions. However, a key challenge remains in verifying the accuracy and reliability of LLM-generated outputs. In particular, the issue of hallucination—where models produce factually incorrect or misleading descriptions—poses a significant limitation for real-world deployment. Future research could aim to improve the robustness and interpretability of LLM-generated annotations, ensuring their reliability for practical use in disaster assessment and emergency response applications.

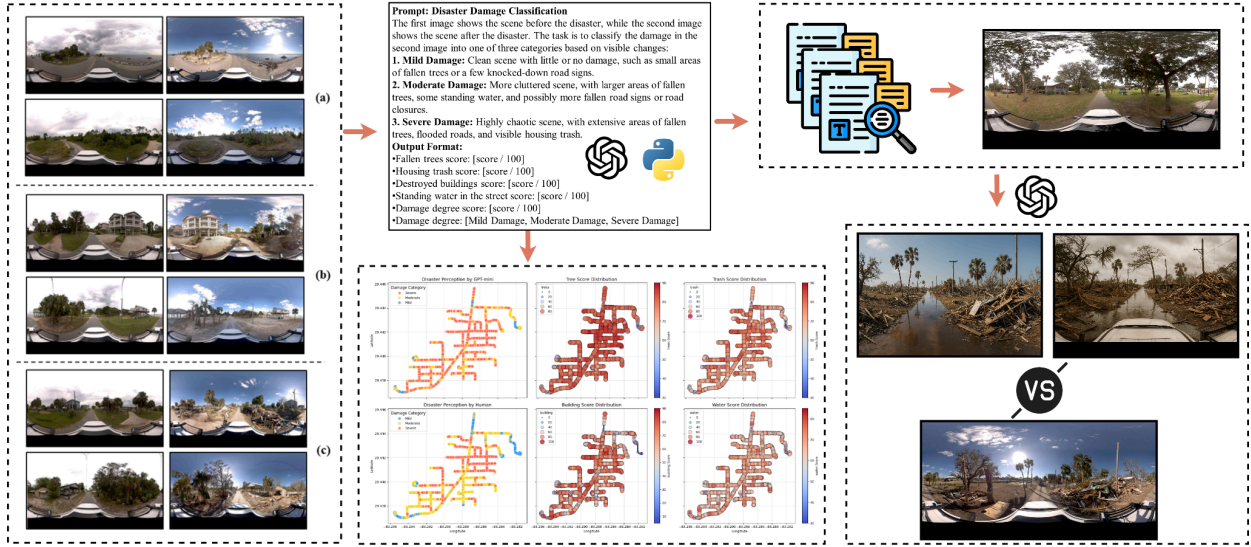

Figure 9. Exploring pre-disaster and post-disaster imagery using ChatGPT.

Another important direction is the integration of multimodal data to improve damage estimation. Current disaster perception models primarily rely on one data source, e.g., street-view imagery or remote sensing data. Nevertheless, combining multiple modalities—such as textual descriptions obtained from social and news media, street-view imagery, and remote sensing data—could significantly enhance classification accuracy and interpretability. Multimodal learning allows models to leverage complementary information from different data sources, improving the robustness of damage estimation. Additionally, integrating structured geographic and environmental data, such as land use maps and real-time meteorological data, could further refine disaster damage predictions. Future research could focus on developing effective fusion techniques to seamlessly integrate diverse data sources.

Lastly, enriching temporal information by incorporating multi-temporal street-view imagery and video data presents another promising avenue for improvement. A key limitation of this study is the temporal inadequacy of pre-disaster imagery, as the available pre-disaster street-view images were captured in late August 2023, which may not fully represent conditions immediately before the hurricane landfall. Ideally, disaster-prone areas could be systematically monitored with planned pre-disaster image collection, ensuring that the most recent baseline data are available. Furthermore, post-disaster image collection could extend beyond the immediate aftermath to cover different recovery phases, allowing for a more comprehensive disaster impact analysis. Incorporating sequentially captured street-view imagery or video data could provide additional insights into how disaster damage evolves over time, enabling models to capture the progression of destruction and restoration rather than treating each image as an independent snapshot. This temporal dimension would enhance disaster monitoring capabilities and improve predictive modeling for future events.

## 6 Conclusion

This study explores the potential of bi-temporal street-view imagery for hyperlocal disaster damage assessment, using data collected before and after the 2024 Hurricane Milton in Horseshoe Beach, Florida. We constructed a dataset containing 2,249 pairs of pre- and post-disaster street-view images and manually labeled them into three damage categories: mild, moderate, and severe. Based on this dataset, we designed three experiments to evaluate pre-trained vision models: (1) single-channel classification using post-disaster imagery, (2) four-class classification incorporating pre-disaster images as a "no damage" category, and (3) a dual-channel architecture that explicitly models differences between pre- and post-disaster image pairs. We implemented seven dual-channel models, leveraging advanced pre-trained vision backbones (ConvNeXt and Swin Transformer), with strategies including simple concatenation, feature fusion, cross-attention, and Siamese difference computation. Model performance was evaluated using weighted-average precision, recall, F1 score, and category-specific recall across the three damage levels.

Our results show that incorporating pre-disaster imagery and employing the designed dual-channel framework significantly improve classification performance. Feature-Fusion ConvNeXt achieves the highest performance among all models, with a validation accuracy of 77.11%, far exceeding the baseline model. Cross-attention-based models also show eminent performance, especially for categorizing images into moderate (e.g., Dual-Swin + Cross-Attention: 91.97%) or severe (e.g., Dual-ConvNeXt+ Cross-Attention: 91.37%). In contrast, the Swin Transformer model that only uses post-disaster street view imagery as input has limited ability to distinguish subtle damages. Grad-CAM visualization confirms that the more accurate models can more effectively focus on damage-related areas, such as collapsed structures and areas filled with debris. These models generally perform better on the moderate and severe damage categories. These findings emphasize the value of bi-temporal imagery and dual-channel architectures in disaster analysis, offering a path toward more accurate, interpretable, and scalable damage assessments. The dataset and methods developed in this study provide a foundation for future work in multi-temporal disaster monitoring and shed light on developing early warning systems and post-disaster recovery planning tools.

In the future, we plan to explore the integration of LLMs to automate the annotation of post-disaster street-view imagery. By incorporating these textual descriptions into multi-modal frameworks, we aim to enhance the ability of AI-based models in interpreting visual content from both semantic and contextual perspectives. This approach has the potential to reduce the burden of manual labeling, improve dataset scalability, and provide more interpretable insights into disaster impacts. In particular, combining language and vision modalities could support more nuanced assessments of structural damage, environmental change, and scene complexity in disaster-affected areas. This line of research opens new possibilities for developing intelligent, scalable management systems for disaster monitoring, reporting, and situational awareness.

**Acknowledgements**

We would like to acknowledge the funding agencies for sponsoring this investigation. This article is based on work supported by two grants. One is from the U.S. National Science Foundation - Collaborative Research: HNDS-I: Cyberinfrastructure for Human Dynamics and Resilience Research (Award No.

2318206). The other grant is from the Gulf Research Program under the U.S. National Academies of Sciences, Engineering, and Medicine (SCON-10000653). Any opinions, findings, conclusions, or recommendations expressed in this material are those of the authors and do not necessarily reflect the views of the funding agencies.

**Disclosure statement**

No potential conflict of interest was reported by the authors.

**References**

Asokan, A., Anitha, J., Patrut, B., Danciulescu, D., & Jude Hemanth, D. (2020). Deep Feature Extraction and Feature Fusion for Bi-temporal Satellite Image Classification. *Computers, Materials & Continua*, *66*(1), 373–388. https://doi.org/10.32604/cmc.2020.012364

de Alwis Pitts, D. A., & So, E. (2017). Enhanced change detection index for disaster response, recovery assessment and monitoring of accessibility and open spaces (camp sites). *International Journal of Applied Earth Observation and Geoinformation*, *57*, 49–60. https://doi.org/10.1016/j.jag.2016.12.004

Gao, S., Zhao, H., Chen, J., Gan, C., Huang, G., & Long, Y. (2025). Deciphering physical disorder of urban street space in China's rust belt: Identification, perception, and interpretation through street-view images. *URBAN DESIGN International*. https://doi.org/10.1057/s41289-024-00264-1

Han, X., Zhang, Z., Ding, N., Gu, Y., Liu, X., Huo, Y., Qiu, J., Yao, Y., Zhang, A., Zhang, L., Han, W., Huang, M., Jin, Q., Lan, Y., Liu, Y., Liu, Z., Lu, Z., Qiu, X., Song, R., … Zhu, J. (2021). Pre-trained models: Past, present and future. *AI Open*, *2*, 225–250. https://doi.org/10.1016/j.aiopen.2021.08.002

Ito, K., Kang, Y., Zhang, Y., Zhang, F., & Biljecki, F. (2024). Understanding urban perception with visual data: A systematic review. *Cities*, *152*, 105169. https://doi.org/10.1016/j.cities.2024.105169

Koshy, R., & Elango, S. (2023). Multimodal tweet classification in disaster response systems using transformer-based bidirectional attention model. *Neural Computing and Applications*, *35*(2), 1607–1627. https://doi.org/10.1007/s00521-022-07790-5

Li, H., Deuser, F., Yina, W., Luo, X., Walther, P., Mai, G., Huang, W., & Werner, M. (2024). *Cross-View*

*Geolocalization and Disaster Mapping with Street-View and VHR Satellite Imagery: A Case Study of Hurricane IAN* (No. arXiv:2408.06761). arXiv. https://doi.org/10.48550/arXiv.2408.06761

Li, L., Ye, Y., Jiang, B., & Zeng, W. (2024, June 6). *GeoReasoner: Geo-localization with Reasoning in Street Views using a Large Vision-Language Model*. Forty-first International Conference on Machine Learning. https://openreview.net/forum?id=WWo9G5zyh0

Liu, F., An, J., Liu, J., Yang, J., Tang, X., & Xiao, L. (2024). Conjoint Cross-Attention Modeling and Joint Feature Calibrating for Remote Sensing Image Change Detection via a Triple-Double Network. *IEEE Transactions on Geoscience and Remote Sensing*, *62*, 1–16. IEEE Transactions on Geoscience and Remote Sensing. https://doi.org/10.1109/TGRS.2024.3393422

Liu, Z., Hu, H., Lin, Y., Yao, Z., Xie, Z., Wei, Y., Ning, J., Cao, Y., Zhang, Z., Dong, L., Wei, F., & Guo, B. (2022). *Swin Transformer V2: Scaling Up Capacity and Resolution*. 12009–12019. https://openaccess.thecvf.com/content/CVPR2022/html/Liu_Swin_Transformer_V2_Scaling_Up_Capacity_and_Resolution_CVPR_2022_paper.html

Liu, Z., Lin, Y., Cao, Y., Hu, H., Wei, Y., Zhang, Z., Lin, S., & Guo, B. (2021). *Swin Transformer: Hierarchical Vision Transformer Using Shifted Windows*. 10012–10022. https://openaccess.thecvf.com/content/ICCV2021/html/Liu_Swin_Transformer_Hierarchical_Vision_Transformer_Using_Shifted_Windows_ICCV_2021_paper

Luo, J., Zhou, F., Yang, J., & Xing, M. (2023). DAFCNN: A Dual-Channel Feature Extraction and Attention Feature Fusion Convolution Neural Network for SAR Image and MS Image Fusion. *Remote Sensing*, *15*(12), Article 12. https://doi.org/10.3390/rs15123091

Mabon, L. (2016). Charting Disaster Recovery via Google Street View: A Social Science Perspective on Challenges Raised by the Fukushima Nuclear Disaster. *International Journal of Disaster Risk Science*, *7*(2), 175–185. https://doi.org/10.1007/s13753-016-0087-4

Mahabir, R., Schuchard, R., Crooks, A., Croitoru, A., & Stefanidis, A. (2020). Crowdsourcing Street View Imagery: A Comparison of Mapillary and OpenStreetCam. *ISPRS International Journal of

*Geo-Information*, *9*(6), Article 6. https://doi.org/10.3390/ijgi9060341

Maswadi, K., Alhazmi, A., Alshanketi, F., & Eke, C. I. (2024). The empirical study of tweet classification system for disaster response using shallow and deep learning models. *Journal of Ambient Intelligence and Humanized Computing*, *15*(9), 3303–3316. https://doi.org/10.1007/s12652-024-04807-w

Neuhold, G., Ollmann, T., Rota Bulo, S., & Kontschieder, P. (2017). *The Mapillary Vistas Dataset for Semantic Understanding of Street Scenes*. 4990–4999. https://openaccess.thecvf.com/content_iccv_2017/html/Neuhold_The_Mapillary_Vistas_ICCV_2017_paper.html

Pavani, T. D. N., & Malla, S. (2024). A review of deep learning techniques for disaster management in social media: Trends and challenges. *The European Physical Journal Special Topics*. https://doi.org/10.1140/epjs/s11734-024-01172-9

Quan, D., Tang, Z., Wang, X., Zhai, W., & Qu, C. (2022). LPI Radar Signal Recognition Based on Dual-Channel CNN and Feature Fusion. *Symmetry*, *14*(3), Article 3. https://doi.org/10.3390/sym14030570

Sun, H., Xu, H., He, H., Wei, Q., Yan, Y., Chen, Z., Li, X., Zheng, J., & Li, T. (2023). A Spatial Analysis of Urban Streets under Deep Learning Based on Street View Imagery: Quantifying Perceptual and Elemental Perceptual Relationships. *Sustainability*, *15*(20), Article 20. https://doi.org/10.3390/su152014798

Smith, A. B. (2020). *U.S. Billion-dollar Weather and Climate Disasters, 1980—Present (NCEI Accession 0209268)* [Dataset]. NOAA National Centers for Environmental Information. https://doi.org/10.25921/STKW-7W73

Tao, W., Yan, X., Wang, Y., & Wei, M. (2024). MFFDNet: Single Image Deraining via Dual-Channel Mixed Feature Fusion. *IEEE Transactions on Instrumentation and Measurement*, *73*, 1–13. IEEE Transactions on Instrumentation and Measurement. https://doi.org/10.1109/TIM.2023.3346498

Wang, C., Antos, S. E., Gosling-Goldsmith, J. G., Triveno, L. M., Zhu, C., von Meding, J., & Ye, X.

(2024). Assessing Climate Disaster Vulnerability in Peru and Colombia Using Street View Imagery: A Pilot Study. *Buildings*, *14*(1), Article 1. https://doi.org/10.3390/buildings14010014

Wang, L., Fang, Y., Li, Z., Wu, C., Xu, M., & Shao, M. (2024). Summator–Subtractor Network: Modeling Spatial and Channel Differences for Change Detection. *IEEE Transactions on Geoscience and Remote Sensing*, *62*, 1–12. IEEE Transactions on Geoscience and Remote Sensing. https://doi.org/10.1109/TGRS.2024.3349638

Woo, S., Debnath, S., Hu, R., Chen, X., Liu, Z., Kweon, I. S., & Xie, S. (2023). *ConvNeXt V2: Co-Designing and Scaling ConvNets With Masked Autoencoders*. 16133–16142. https://openaccess.thecvf.com/content/CVPR2023/html/Woo_ConvNeXt_V2_Co-Designing_and_Scaling_ConvNets_With_Masked_Autoencoders_CVPR_2023_paper.html

Xie, Z., Lin, Y., Yao, Z., Zhang, Z., Dai, Q., Cao, Y., & Hu, H. (2021). *Self-Supervised Learning with Swin Transformers* (No. arXiv:2105.04553). arXiv. https://doi.org/10.48550/arXiv.2105.04553

Xue, Z., Zhang, X., Prevatt, D. O., Bridge, J., Xu, S., & Zhao, X. (2024). *Post-hurricane building damage assessment using street-view imagery and structured data: A multi-modal deep learning approach* (No. arXiv:2404.07399). arXiv. https://doi.org/10.48550/arXiv.2404.07399

Yang, Y., Wang, S., Li, D., Sun, S., & Wu, Q. (2024). GeoLocator: A Location-Integrated Large Multimodal Model (LMM) for Inferring Geo-Privacy. *Applied Sciences*, *14*(16), Article 16. https://doi.org/10.3390/app14167091

Yu, W., Zhou, P., Yan, S., & Wang, X. (2024). *InceptionNeXt: When Inception Meets ConvNeXt*. 5672–5683. https://openaccess.thecvf.com/content/CVPR2024/html/Yu_InceptionNeXt_When_Inception_Meets_ConvNeXt_CVPR_2024_paper.html

Zhai, W., & Peng, Z.-R. (2020). Damage assessment using Google Street View: Evidence from Hurricane Michael in Mexico Beach, Florida. *Applied Geography*, *123*, 102252. https://doi.org/10.1016/j.apgeog.2020.102252

Zhang, D., Yang, J., Li, F., Han, S., Qin, L., & Li, Q. (2022). Landslide Risk Prediction Model Using an

Attention-Based Temporal Convolutional Network Connected to a Recurrent Neural Network. *IEEE Access*, *10*, 37635–37645. IEEE Access. https://doi.org/10.1109/ACCESS.2022.3165051

Zhao, X., Lu, Y., & Lin, G. (2024). An integrated deep learning approach for assessing the visual qualities of built environments utilizing street view images. *Engineering Applications of Artificial Intelligence*, *130*, 107805. https://doi.org/10.1016/j.engappai.2023.107805

Zhao, X., Ma, H., Niu, C., Gao, L., & Kong, X. (2024). *Landslide susceptibility assessment based on ConvNext in Longyang district of Baoshan city, China*. Research Square. https://doi.org/10.21203/rs.3.rs-4897122/v1

Zheng, J. (2024). ConvNeXt-Mask2Former: A Semantic Segmentation Model for Land Classification in Remote Sensing Images. *2024 5th International Conference on Computer Vision, Image and Deep Learning (CVIDL)*, 676–682. https://doi.org/10.1109/CVIDL62147.2024.10603728

Zong, Y., Bohdal, O., Yu, T., Yang, Y., & Hospedales, T. (2024). *Safety Fine-Tuning at (Almost) No Cost: A Baseline for Vision Large Language Models* (No. arXiv:2402.02207). arXiv. https://doi.org/10.48550/arXiv.2402.02207